# Mary Astell's words in *A Serious Proposal to the Ladies* (part I), a lexicographic inquiry with NooJ

**Hélène Pignot**
*SAMM, University of Panthéon-Sorbonne*
*Paris, France.*

**Odile Piton**
*SAMM, University of Panthéon-Sorbonne*
*Paris, France.*

## Abstract

In the following article we elected to study with NooJ the lexis of a 17th century text, Mary Astell's seminal essay, *A Serious Proposal to the Ladies*, part I, published in 1694. We first focused on the semantics to see how Astell builds her vindication of the female sex, which words she uses to sensitise women to their alienated condition and promote their education. Then we studied the morphology of the lexemes (which is different from contemporary English) used by the author, thanks to the NooJ tools we have devised for this purpose. NooJ has great functionalities for lexicographic work. Its commands and graphs prove to be most efficient in the spotting of archaic words or variants in spelling.

## Introduction

In our previous articles, we have studied the singularities of 17th century English within the framework of a diachronic analysis thanks to syntactical and morphological graphs and thanks to the dictionaries we have compiled from a corpus that may be expanded overtime. Our early work was based on a limited corpus of English travel literature to Greece in the 17th century. This article deals with a late seventeenth century text written by a woman philosopher and essayist, Mary Astell (1666–1731), considered as one of the first English feminists. Astell wrote her essay at a time in English history when women were "the weaker vessel" and their main business in life was to charm and please men by their looks and submissiveness. In this essay we will see how NooJ can help us analyse Astell's rhetoric (what point of view does she adopt, does she speak in her own name, in the name of all women, what is her representation of men and women and their relationships in the text, what are the goals of education?). Then we will turn our attention to the morphology of words in the text and use NooJ commands and graphs to carry out a lexicographic inquiry into Astell's lexemes.

## I. About the Author and Remarks on the Text

### *1. 1. Who was Mary Astell?*

In *A Serious Proposal to the Ladies*, part I, first published anonymously in 1694, Mary Astell makes a rather odd suggestion (for the period!): women should withdraw from the world in a "blest abode" that she even dares to call a convent (in a country where all convents had been closed down). There they might enjoy each other's company, spend their time in study, good works and prayer (without forgetting daily church attendance). This place could also be a refuge for heiresses and unmarried women who need to escape the assiduities of adventurers and fortune seekers. Her book was widely discussed in her day and satirised by the moralist Richard Steele in the *Tatler* in 1709, calling Astell the leader of "an order of Platonick Ladies" bent on celibacy and "resolv'd to join their Fortunes and erect a Nunnery" in Steele (1709).
According to the historian Antonia Fraser (1984, 404), what inspired Astell to write her proposal was the life of her friend Hortense Mancini, Duchesse de Mazarin, King Charles

II's ex-mistress, who was the subject of numerous scandals, such as "the running away in Disguise with a Spruce Cavalier"! Astell was convinced her friend's "unhappy shipwreck" pointed out "the dangers of an ill Education and unequal marriage" (1700, 3). *A Serious Proposal to the Ladies*, part I, is signed "by a Lover of her Sex" and was first published anonymously in 1694 when the author was only 28. Part II (which will not be tackled here) is an application of Descartes' and Malebranche's philosophy to the education of women.

### *1. 2. From the Text to the Digital Text*

OCR software could not be used for this text so we had to type the text preserving the original spelling and punctuation. NooJ does not keep the italics, hence the literary effects created by Astell's use of italics are lost. In 17th century printing, most nouns are capitalised. The last word (or its last syllable) on the page is repeated at the top of the next page: here "ous" and "home" in illustration 1.

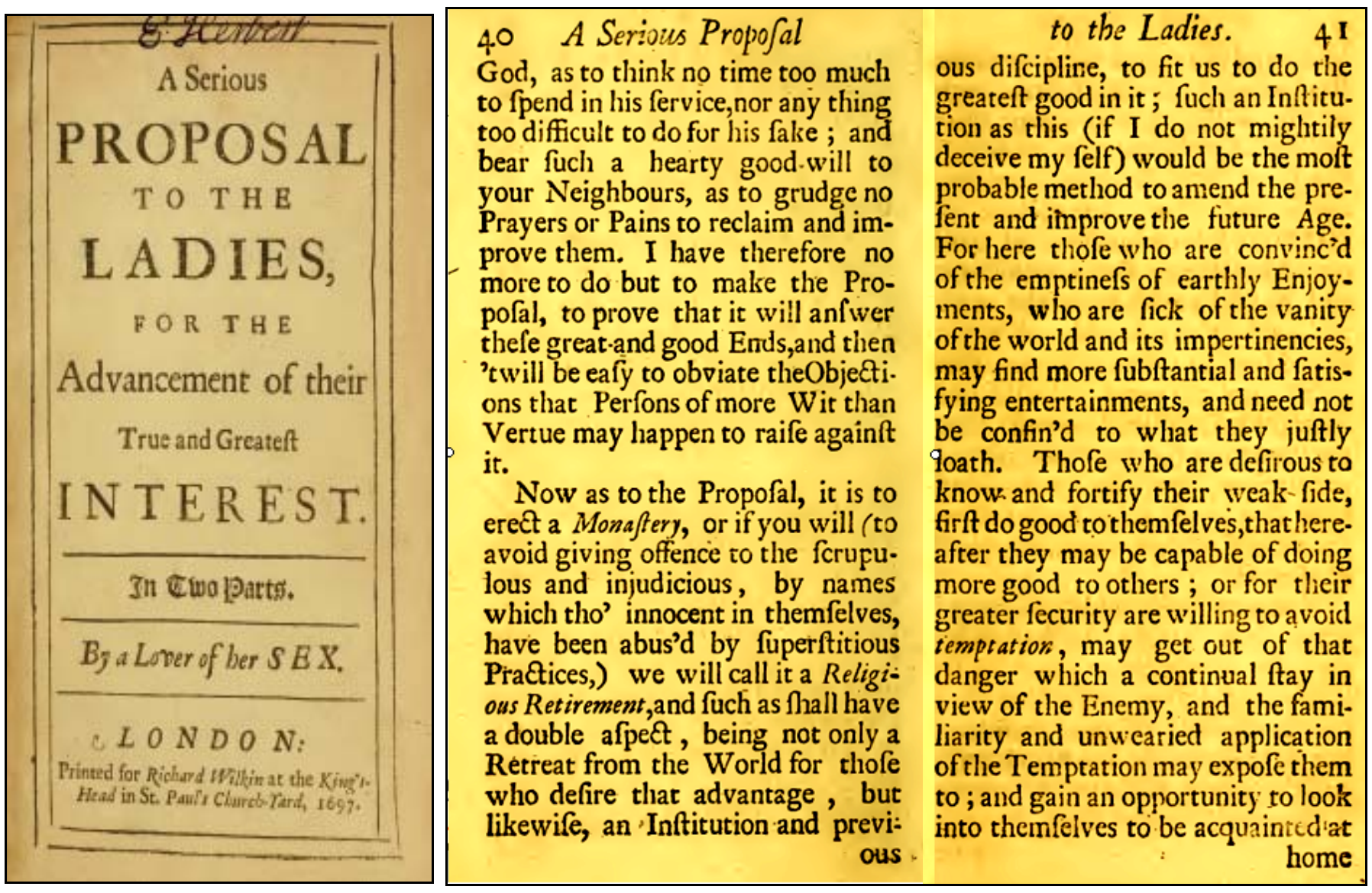
A Serious
PROPOSAL
TO THE
LADIES,
FOR THE
Advancement of their
True and Greateſt
INTEREST.
In Two Parts.
By a Lover of her SEX.
LONDON:
Printed for Richard Wilkin at the King's-Head in St. Paul's Church-Yard, 1697.

40 A Serious Propoſal

God, as to think no time too much to ſpend in his ſervice,nor any thing too difficult to do for his ſake ; and bear ſuch a hearty good-will to your Neighbours, as to grudge no Prayers or Pains to reclaim and improve them. I have therefore no more to do but to make the Propoſal, to prove that it will anſwer theſe great and good Ends,and then 'twill be eaſy to obviate theObjections that Perſons of more Wit than Vertue may happen to raiſe againſt it.

Now as to the Propoſal, it is to erect a *Monaſtery*, or if you will (to avoid giving offence to the ſcrupulous and injudicious, by names which tho' innocent in themſelves, have been abus'd by ſuperſtitious Practices,) we will call it a *Religious Retirement*,and ſuch as ſhall have a double aſpect, being not only a Retreat from the World for thoſe who deſire that advantage, but likewiſe, an Inſtitution and previous

to the Ladies. 41

ous diſcipline, to fit us to do the greateſt good in it ; ſuch an Inſtitution as this (if I do not mightily deceive my ſelf) would be the moſt probable method to amend the preſent and improve the future Age. For here thoſe who are convinc'd of the emptineſs of earthly Enjoyments, who are ſick of the vanity of the world and its impertinencies, may find more ſubſtantial and ſatisfying entertainments, and need not be confin'd to what they juſtly loath. Thoſe who are deſirous to know and fortify their weak ſide, firſt do good to themſelves,that hereafter they may be capable of doing more good to others ; or for their greater ſecurity are willing to avoid *temptation*, may get out of that danger which a continual ſtay in view of the Enemy, and the familiarity and unwearied application of the Temptation may expoſe them to ; and gain an opportunity to look into themſelves to be acquainted at home

**Illustration 1**: Frontispiece and Excerpt from *A Serious Proposal to the Ladies*

### *1.3. Astell's point of view*

We first focused on the semantics, to examine how Astell builds her vindication of the female sex, which words she uses to sensitise women to their alienated condition and promote their education. As the object of this study is a literary text, it is interesting first to study point of view, which can be examined by focusing on the use of personal pronouns and possessive adjectives in Astell's text.

#### *1.3.1. Personal pronouns.*

The study of personal pronouns highlight different aspects of Astell's rhetoric: she exhorts with the personal pronoun "you", denounces with "they", creates complicity with we, and seldom uses the first person "I".

The hortatory aspect of the text may be highlighted when we extract the sentences in which the pronoun “you” is used (locate pattern you): “How can you be content to be in the World like Tulips in a Garden, to make a fine shew and be good for nothing; have all your Glories set in the grave...” This functionality enabled us to extract expressions such as “I pray you”, “I entreat you”, “render you”, “persuade you”, I “would have you”, “all that is required of you”. The text is a protrepsis, an exhortation to women to change their lives and shake off the yoke of ignorance and superficiality.

The frequent use of the pronoun “we” is meant to encourage identification with the author and with a community of women sharing the same fate in a male-dominated society.

By contrast, the occurrences of “I, me, my” account for 4.56 % of pronouns. Astell does not use the pronouns “I” and “me” very much: these self-references (84) account for only 0.44% of the words of the text, which points to her self-effacement as an author, to be linked with that of the female author in the 17th century. Incidentally, it should be noted that her book is signed “by a Lover of her Sex”, and not under her own name. In her preface to *Letters Concerning the Love of God* (1695,4), Astell says she wants “to slide gently through the world without so much as being seen or taken notice of.”

| Personal pronouns | occurrences | Total by group |
|---|---|---|
| Me/ my self / my / I | 12 / 5 / 15 / 52 | 84 |
| Thou / Thy / thine | 0 | 0 |
| She / her | 129 / 183 | 129 |
| He / him / his | 18 / 24 / 24 | 631 |
| We / us / our | 141 / 88 / 144 | 373 |
| You / your | 125 / 81 | 206 |
| They / them / their | 148 / 111 / 159 | 418 |
| Total | 1841 | |

**Table 1**. Number of personal pronouns.

Such self-effacement was indeed common at the time as writing was not considered as an occupation befitting the female sex! It was only in the late 18th century that women were to establish themselves as authors and make a living as novelists. A talented novelist like Sarah Fielding, the sister of Henry Fielding, considered as one of the first 18th century female novelists, published all of her novels anonymously, and lived most of her life in strained circumstances. Indeed, she depended on the generosities of noble patrons who enabled her to publish her books by subscription. As the literary critic Vivien Jones points out: “*for women to write and publish at all was by definition a transgressive and potentially liberating act, a penetration of the forbidden public sphere, and the virulence with which fiction was attacked as a corrupting ‘female’ genre is telling evidence of its disruptive potential*” (1990, 12).

### 1.3.2. References to both sexes in the text

Now let us direct our attention to Astell’s words. Is there anything about them that is provocative or subversive of traditional female roles and women’s relationships with men? Is there any such thing as a feminine nature?

When we established the concordances for “man” and “woman”, “husband” and “wife”, masculine v. feminine and for the pronouns “they”,“we” and “you”, we could extract key sentences like these ones, which show that all the supposed faults or vices that are attributed to women are not part of their nature, but result from a deficient education:

> *"if from our Infancy we are nurs'd up in Ignorance and Vanity; are taught to be Proud and Petulant, Delicate and Fantastick, Humorous and Inconstant, 'tis not strange that the ill effects of this conduct appear in all the future actions of our lives."*

> *"that Ignorance is the cause of most Feminine Vices, may be instanc'd in that Pride and Vanity which is usually imputed to us, and which I suppose -if thoroughly sifted, will appear to be some way or other, the rise and Original of all the rest. These, tho' very bad Weeds, are the product of a good Soil, they are nothing else but Generosity degenerated and corrupted. A desire to advance and perfect its Being, is planted by GOD in all Rational Natures, to excite them hereby to every worthy and becoming Action."*

Astell likes to use vegetal metaphors to make her point: uneducated women are compared to "*tulips in a garden*"[1] and education is likened to gardening. Women are rational (and not sentimental) creatures just like men: "*We value them too much, and our selves too little, if we place any part of our worth in their opinion; and do not think our selves capable of nobler Things than the pitiful Conquest of some worthless heart*"[2] (*Proposal*, 11) and they are trapped in the benign neglect in which men have purposely kept them: "*We're indeed oblig'd to them for their management...So that instead of inquiring why all Women are not wise and good, we have reason to wonder that there are any so. Were the men as much neglected, and as little care taken to cultivate and improve them, perhaps they wou'd be so far from surpassing those whom they now dispise, that they themselves wou'd sink into the greatest stupidity and brutality*" (*Proposal*, 15).

Women and men by custom are made to live in two separate spheres, respectively appearance and essence. Women are made by men creatures of appearance[3], frivolity and thoughtlessness, thanks to education they will have access to the realm of being and knowledge as God has given them –just as men— the faculty of reason: *"The Ladies, I'm sure, have no reason to dislike this Proposal, but I know not how the men will resent [=feel] it, to have their enclosure broken down, and Women invited to tast of that tree of Knowledge they have so long unjustly monopoliz'd. But they must excuse me, if I be as partial to my own Sex as they are to theirs, and think Women as capable of Learning as men are, and that it becomes them as well"* (*Proposal*, 58). Incidentally, a clever, well-educated woman might also conveniently help conceal her companion's intellectual limitations[4]!

### 1.3.3. Astell's favourite words: the goals of women's education

In Astell's text (aside from grammatical words which are not significant here), the most commonly used lexemes are GOD (capitalized most of the time) and world. This may reflect a tension in the text between two conflicting demands for women in the 17th century: finding a place in the world but also accomplishing their spiritual destiny as God's creatures, endowed with reason just as males. The other most frequently used words in Astell's text are soul, virtue, nature, mind, knowledge and piety: these two notions are asserted throughout the text as the goals that women should pursue.

---

[1] "How can you be content to be in the World like Tulips in a Garden, to make a fine shew and be good for nothing" (Proposal, 9).

[2] And further down: "But I will not pretend to correct their Errors, who either are, or at least think themselves too wise to receive Instruction from a Womans Pen" .

[3] At one point in the text she calls dressing "that grand devourer" of time and energy, a remark mothers of teenage girls may still make today!

[4] "The only danger is that the Wife be more knowing than the Husband; but if she be 'tis his own fault, since he wants no opportunities of improvement; unless he be a natural Block-head, and then such an one will need a wise Woman to govern him, whose prudence will conceal it from publick Observation, and at once both cover and supply his defects."

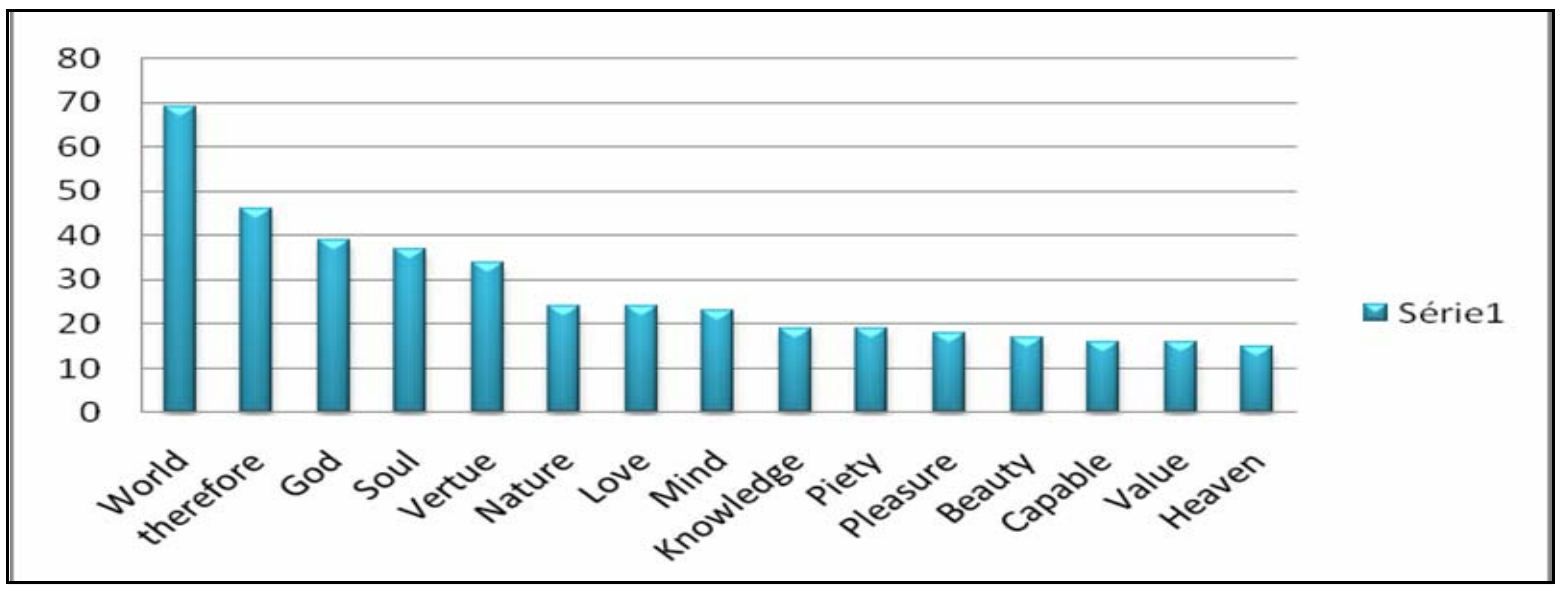

**Table 2.** Astell's 15 favourite lexemes

However, such lists and figures (even they reveal interesting frequent occurrences of certain words) can help literary analysis but not replace it. The absence of a word does not prove that the concept is not in the text; the writer may also use negation, circumlocution or euphemism. The words need to be put back into context (which can be shown by NooJ), as writers can resort to irony or humour, using words and meaning their exact opposites. The extraction of certain sentences from Astell's text highlights the author's frequent use of humour and irony.
Moreover, it must be pointed out that these listings do not provide any information on the sense relationships between individual lexemes. We are also aware that we have not studied collocations and frozen expressions in the text. However, within the scope of our work (recording the singularities of 17th century English), NooJ tools prove to be most efficient when the reader wants to study affixes in a 17th century text, as they help him locate archaic words or words whose prefixation, suffixation and spelling have changed over time.

## II. Morphological study of Astell's words with NooJ

After studying the length of words, we will record the morphological modifications we have observed and described with NooJ. Some lexemes are modified thanks to a punctuation mark such as an apostrophe (marking elision or contraction) or a hyphen. Other modifications include variations in affixes.

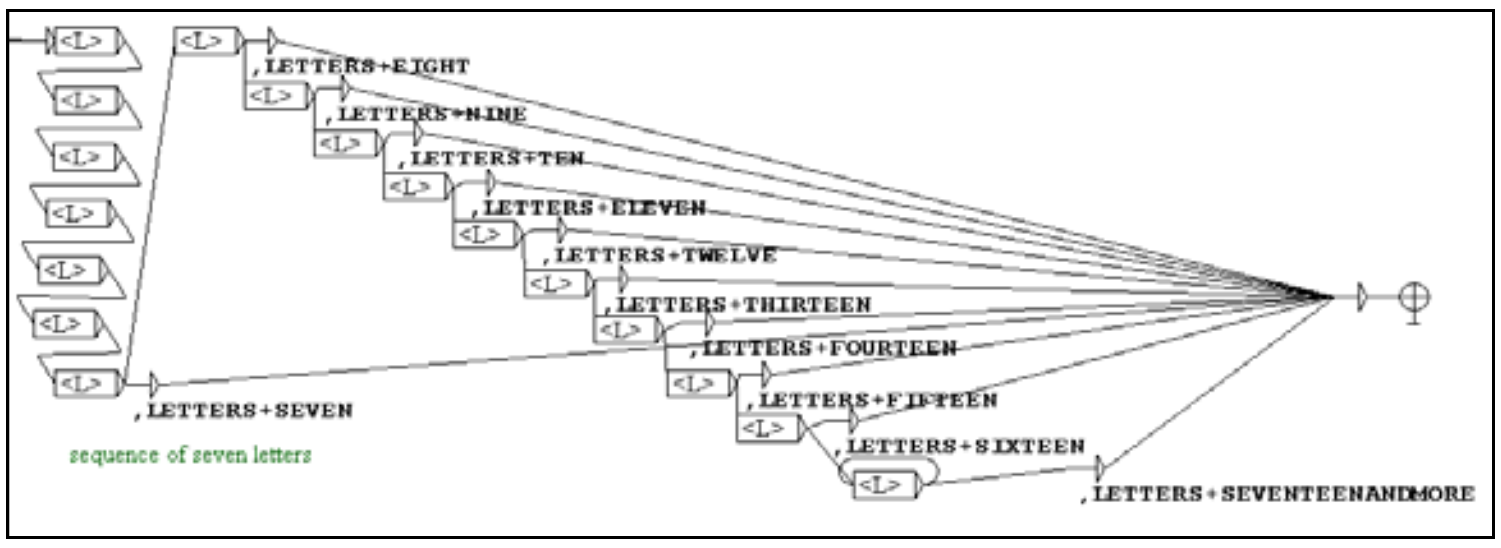

**Illustration 2**: Morphological graph to locate long words

### *2.1 Long words in Astell's text.*

In English long words (as opposed to short Saxon words) are often learned words derived from Latin or Greek. A large number of Latin and Greek words were imported into the English language during the Renaissance. Astell's text is a literary work and we wondered if we might find the percentage of long words in her text thanks to NooJ. To do so, we created a graph that enables NooJ to locate them. We determined their number and the occurrences of each lexeme in a very simple manner. The graph below can tag the words in the text according to their length. Then to look for the occurrence of a N-letter word, we use the command <LETTERS+NUMBER>. Our results are presented in table 3:

| LENGTH | WORDS | OCCURRENCES | OCC per WORD |
|---|---|---|---|
| SEVEN | 515 | 1213 | 2.36 |
| EIGHT | 408 | 819 | 2.01 |
| NINE | 334 | 691 | 2.07 |
| TEN | 251 | 455 | 1.81 |
| ELEVEN | 144 | 248 | 1.73 |
| TWELVE | 79 | 121 | 1.53 |
| THIRTEEN | 37 | 68 | 1.84 |
| FOURTEEN | 16 | 27 | 1.69 |
| FIFTEEN | 10 | 15 | 1.5 |
| SIXTEEN | 1 | 1 | 1 |
| Total | 1795 | 3659 | 2.03 |

**Table 3:** Long words>6 letters

The total number of long words is 3659 out of 18,759 words in the text, i.e. a percentage of around 19.5%. This graph enables us to determine the percentage of long words and spot very long words (e.g. 16: uncharitableness) and some archaic words (e.g. 15: pragmaticalness). On average long words are used about twice (2.03).

## 2.2. Study of compounds and juxtaposed words

### *2.2.1 Hyphenated compounds*

Many words were broken off and spelled as hyphenated compounds, instead of one single word: for instance fore-heads (17th century form) which is spelled foreheads in modern English. We have applied the following syntactical graph to be able to spot these forms. This graph highlights a systematic tendency to isolate the prefix from the root in 17th century spelling, thanks to the hyphenation (first path).

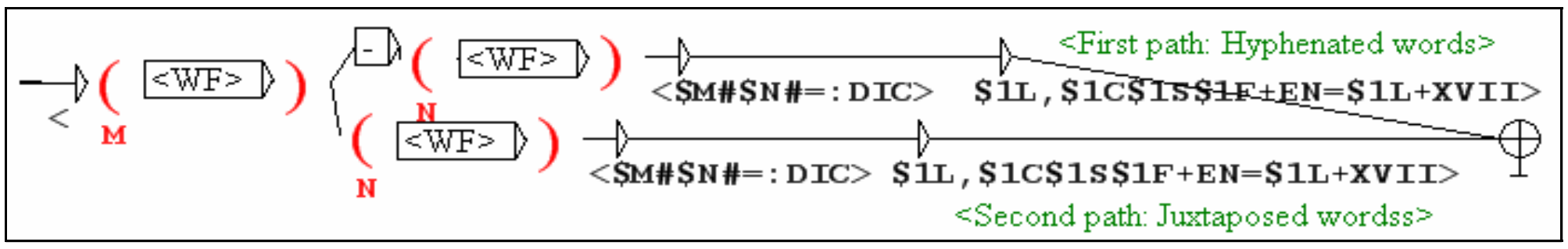

**Illustration 3**: Syntactical graph for compounds

This produced 28 results or occurrences of the form word1-word2 ( for example God-like for godlike or pre-ingage for preengage) recognised by NooJ.

During the search, NooJ also appeals to another graph which enables it to take into account the transformation of the prefix -in into -en or -en into -in (as we saw in a previous article, Piton and Pignot, 2008). Sometimes a fusing together of two hyphenated words may also need to be performed, as in “where-ever” which becomes “wherever”. At our request NooJ was modified by Max Silberstein to permit this transformation.

| NOUNS: 14 | VERBS: 9 | ADJECTIVES: 2 | CONJ or ADV: 2 |
|---|---|---|---|
| Block-head blockhead | me-thinks (impers verb) or methinks (arch): it seems to me | God-like godlike | hence-forward henceforward |
| church-yard churchyard | out-do outdo | over-careful overcareful | where-ever wherever |
| eye-sight eyesight | out-weigh outweigh | | |
| fore-heads foreheads | over-done overdone | | |
| good-will goodwill | over-grown overgrown | | |
| Holy-day holiday | over-rate overrate | | |
| ill-nature illnature | over-run overrun | | |
| non-improvement nonimprovement | over-stock’d overstocked | | |
| non-sense nonsense | pre-ingage preengage (prefix in becomes en) | | |
| out-side outside | | | |
| pre-eminence preeminence | | | |
| set-offs setoffs | | | |
| well-fare welfare | | | |
| well-wishers wellwishers | | | |

**Table 4**. From hyphenated words to concatenated words

### *2.2.2. Juxtaposed words*

Some words result from the concatenation or fusing together of two lexemes, as in the case of personal pronouns (my self for myself). In 17th century English, these words were spelled as juxtaposed words instead of one single entity. We came across a few occurrences of this type of words in Astell's text; the second path in illustration 3 enables us to look for them systematically. Besides personal pronouns such as ‘her self’, ‘our selves’ (whose spelling indicates that “self” was perceived as a noun), we have found 11 compounds: some body, every day, for ever, every one, any one, any thing, often times, like wise, to day, no body, back wardness. However, this link also produces much noise and funny results: *are a,area; be an,bean; direct or,director; not able,notable; yet I, yeti!* Some compounds are still hyphenated in contemporary English, while verbs (except when the prefix ends with the same letter as the first letter of the root) and adjectives no longer isolate the prefix thanks to the hyphenation.

### *2.3 Words with apostrophes*

| **Frequence** | 329 | 2 | 90 | 356 | 1334 |
|---|---|---|---|---|---|
| **Character** | ' | ‘ | ’ | . | , |

**Table 5**: Apostrophes in Astell's text

With a locate pattern of a single ', we have found 421 occurrences of apostrophes in the text. There are more apostrophes than full stops (356) and three times less than commas (1334). In 17th century typography, many words could be spelled with an apostrophe, which marked the elision of a letter, usually the mute e. Besides 118 past participles in ’d or ’t, some ’s and some familiar contractions, we have extracted other contractions that are not in the NooJ dictionary as they are not used in contemporary English. These contractions needed to be adequately described and added to our dictionary. Here are a few examples:

e’er, <ever,ADV> +UNAMB and 'twou'd, <it,it,PRO+3+n+s> <would,V+PT+3+s> +UNAMB. Each entry describes the individual units in the contraction and we have tagged the entry as unambiguous (+UNAMB) when only one interpretation is possible. In the case of tho’t; the apostrophe is used to mark two elisions (an apocope and an aphaeresis): tho’= though and ‘t = it. Our dictionary (which includes Max’ Silberstein's 49 forms and ours) of contracted or elided forms has 90 entries. Many of these contractions are no longer used except in poetry (ex: cou'd : could; e'er: ever; ev'ry: every; heav'n: heaven). Any word containing an e mute could be elided in 17$^{th}$ century spelling (a graph could be made for this).

### 2.4 A Study of affixes in Astell's text

#### 2.4.1 Inflectional affixes for adjectives, adverbs and verbs

In 17$^{th}$ century English any adverb or adjective whatever its length could form the comparative by adding the affix -er and the superlative by adding -est. The following graph may locate superlatives and comparatives; in Astell's text it can locate the archaic form worser (a double comparative of bad).

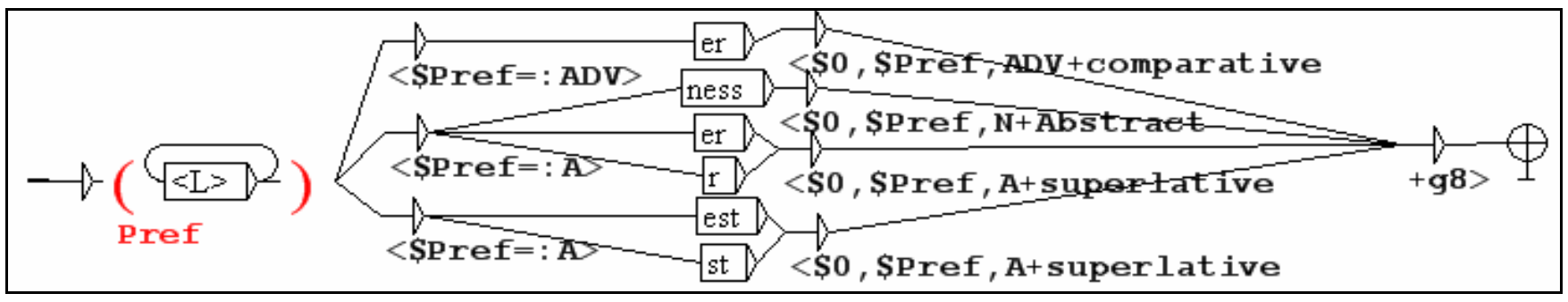

**Illustration 4**: Recognition of the comparative and the superlative

Astell only uses the comparative in -er and the superlative in -est for short words.
Among grammatical affixes that are specific to 17$^{th}$ century English, mention should be made of the verb inflexions in -est and -eth (st, 2$^{nd-}$person singular, -th or -st 3$^{rd}$ person singular) in the present and the preterit.
They may be spotted thanks to the graph below, and are recognised as valid verb forms when s or es is added to the root. It should be noted that there is only one occurrence of this form in our text (“profiteth”) which dates from 1694.

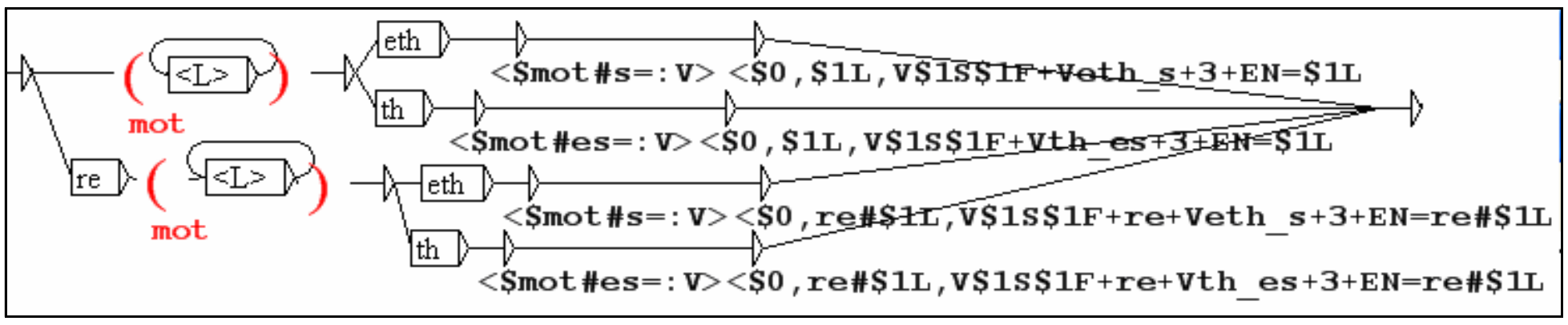

**Illustration 5**: Morphological graph to recognise the third person present

We can also automatically recognise the absence of affix in the subjunctive in clauses introduced by a conjunction like lest (also spelled least): “least she only *change* the instance and *retain* the absurdity”. The subjunctive was in full use as a thought-mood in 17$^{th}$ century English as we saw in a previous article (Piton and Pignot, 2008), not only after certain adjectives and verbs as in contemporary English, but in conditional, concessive and temporal clauses. The infinitive may be used for all persons.

### *2.4.2. variations in affixes*

In a previous article when we tackled our corpus of English travellers to Greece, we noticed variations in affixes (prefixes or suffixes). This new study is different as it simply uses simple existing NooJ commands to list prefixes and suffixes systematically and compare the word forms with their forms in contemporary English and spot the words that are archaic or have fallen into disuse. To do so, we have taken up the classification of affixes described by Tournier (2004). We decided not to specify the grammatical category (noun, verb, adjective) of the word form to make sure we could find all the results and sort them by grammatical category afterwards. We made a general inquiry into all the prefixes and suffixes mentioned by Tournier and only present the most significant results here.

## *Prefixes*

As far as adjectival prefixes are concerned, Tournier's description includes quantitative prefixes (demi, semi, hemi, uni, ono, bi, di, diplo, ter/tri, quadri, quinque, penta, centi, hecto, ulti, pluri, poly), antonymic prefixes (a, in, un, non-, anti, dis) and spatial prefixes (epi, super, circum, peri, hypo, infra, sub, endo, intra). In Astell's text, there are no results for quantitative and spatial adjectives.

To locate prefixes thanks to NooJ, we need to use the following command <WF+MP="^prefix">, which can spot the variation of the prefix from "in" to "en" and vice versa for verbs: for the "en" prefix NooJ found 41 answers, such as encrease (increase in CE); conversely for the in prefix we get 42 answers and five verbs could be found this way such as inforce, ingage, incourage (enforce etc. in contemporary English).

This method also allows us to locate words starting with "dis" with a different spelling or generally speaking long, obsolete words that are no longer used in contemporary English or with a different meaning. For the prefix -dis among 78 answers, it found the verb dispise, a spelling variant of despise in contemporary English which comes from the Latin *despicere*) and nouns such as disquisition (a formal enquiry, this word first appears in 1640 according to the OED), dispensatory (which means pharmacopoeia, a book in which medical substances and their properties are described). As the grammarian James Howell explains (1662, 10), e and i "supply one anothers place" and are used indifferently in English spelling at the time, just as in Spanish and Italian.

## *Suffixes*

Dictionaries and grammars contain lists of prefixes. To locate suffixes in a text with NooJ, we need to use the command "locate pattern": <WF+MP="suffix$">. For NooJ, a suffix is a chain of characters, not a semantic entity. As far as nouns and verbs are concerned we observed variants that we have recapitulated further down. The main noun suffixes that we investigated are shown in table 6. The total number of archaic nouns or archaic forms for nouns located thanks to the study of suffixes is 26.

| Suffix | Number of nouns | Number of occurrences | Nouns with an archaic meaning or spelling |
|---|---|---|---|
| tion/tions | 98/17 | 171/25 | commendation, concoction, disquisition |
| ity/ities | 52/12 | 100/22 | |
| ness/nesses | 49/2 | 78/2 | forwardness, pragmaticalness, back wardness, dearnesses |
| ment/ments | 28/22 | 61/33 | |
| ence/ences | 25/7 | 47/12 | |
| ance/ances | 17/6 | 42/10 | complyance |
| ure/ures | 13/11 | 58/21 | contexture |
| ency/encies | 7/5 | 8/9 | impertinency, indifferency, innocency, subserviency, displacencies |
| sion/sions | 6/6 | 14/13 | propensions |
| er/ers (human) | 6/13 | 11/10 | |
| or/ors | 08/10 | 11/14 | taylor, traytors |
| acy/acies | 6/1 | 6/1 | apostacy |
| ancy/ancies | 4/1 | 5/2 | incogitancy |
| ive/ives | 4/3 | 5/4 | preparative, persuasives |
| ary/aries | 2/1 | 2/1 | |
| ist/ists | 2/0 | 2/0 | |
| al/als | 2/6 | 9/7 | intellectuals, generals, temporals |
| ast/asts | 1/0 | 1/0 | antepast |
| yon/yons | 0/1 | 0/1 | cyons |

**Table 6**: Archaic forms or meanings for nouns

For the suffix 'ness' we get 48 results, among which some archaic words: pragmaticalness (opinionatedness or stubbornness in CE), frowardness (disposition to disobedience and opposition), and one archaic form: back wardness spelled in two words that we added to our dictionary of 17th century English. The most productive suffixes seem to be -tion and -ness.

Here are other obsolete forms or meanings for words with the suffixes -ure, -tion, -cy, and -ce: contexture, disquisition, innocency, indifferency, subserviency, acquiesce, the verb "acquiesce in" having a specific meaning which is "remain at rest". The variation from -ence to -ency concerns nouns and when there are two forms of the same word the competing form in -ency tends to become archaic.

As regards adjectives and verbs, we systematically counted and studied adjectives in ous (117), al (91), able (61), ary (30), ly (26), ish (15), ate (15), less (15), ick (7), ical (6), ic (2), ose (0) and ist (0). We only observed 10 variations in spelling: woful (woeful); fantastick; heroick; flitting (fleeting), pityable (pitiable); changeable; improveable; unblameable; publick; vertuous (virtuous).

For verbs we found thirteen archaic forms: the verbs "disinteress", "attone", "glorifie" and "rectifie", six archaic spellings of the past participle (ex: "deprest, affixt"), and four gerunds: rejoycing (rejoicing), blustring, loosning and mouldring –in these three forms the mute e is elided.

### *2.5. Numerical assessment*

If we do not apply the graphs and dictionaries we have created, NooJ finds 199 unknown words, such as improveable, improveing, incogitancy, incourage, inferiour, inforce, or ingage. The tools we have been developing with NooJ enabled us to correctly process 190 archaic words. After applying our graphs and dictionaries there were still 9 unknown forms (accrew, Aegypt, benumb, buz, cyons, meerly, mormo, shewn, viz) that we have added to our dictionary. In the case of "benumb" and "viz." these are not archaic forms or meanings but words that are missing from the sdic NooJ dictionary of contemporary English. Our systematic study of prefixes and suffixes enabled us to locate 49 archaic forms (nouns, verbs and adjectives).

**Conclusion**

From a semantic viewpoint, NooJ's functionalities (such as locate pattern) allow the reader to study the lexis of a text and single out its main lexical fields and recurrent words. Astell's essay is a very positive and optimistic advocacy of women's education, a proto-feminist text that has potent anti-male overtones as we saw in part I of this presentation. What motivates Astell in writing her text is above all a sincere wish to see women improve themselves thanks to education and fulfill themselves spiritually and intellectually.

When we read the text linearly, we counted 69 words out of 18,734 that have a specific meaning in 17th century English. NooJ could not locate every of them because their forms are in its dictionary, but could have other archaic meanings in 17th century English, for example "to close with a proposal" which means to accept it, "fantastick" which is synonymous with eccentric or quaint, "generals" (generalities), and "temporals" (temporal matters). We have added all these archaic meanings to our NooJ dictionary with a specific presentation which is meant to facilitate their recognition and listing. When the word varies only in spelling, the modern spelling is indicated next to the entry which reads: accrew,accrue,V+EN=accrue+Dic_EN_XVII+spelling.

When the meaning (not the form) of the word is different and archaic, we indicate it: displacency,N+EN=displeasure+Dic_EN_XVII+meaning. Each archaic meaning of a word necessitates a new entry.

All these NooJ tools will help us in a great but arduous work in progress, the creation of a dictionary of 17th century English which we would like to put at the disposal of the NooJ community.

## Primary sources